\documentclass[10pt, conference]{IEEEtran}
\IEEEoverridecommandlockouts
\usepackage{cite}
\usepackage{amsmath,amssymb,amsfonts}
\usepackage{algorithmic}
\usepackage{graphicx}
\usepackage{textcomp}
\usepackage{xcolor}
\usepackage{bm}
\usepackage{booktabs}
\usepackage{gensymb}
\usepackage{subfig}
\usepackage{threeparttable}
\usepackage{float}
\usepackage{stfloats}
\usepackage{url}
\usepackage{comment}

\def\BibTeX{{\rm B\kern-.05em{\sc i\kern-.025em b}\kern-.08em
    T\kern-.1667em\lower.7ex\hbox{E}\kern-.125emX}}

\begin{document}

\title{RadProPoser: Probabilistic Radar Tensor Human Pose Estimation That Knows Its Limits
\thanks{This work was partly funded by the Deutsche Forschungsgemeinschaft
(DFG, German Research Foundation) -- SFB~1483 -- Project-ID~442419336,
\textit{EmpkinS}. This work has been submitted to the IEEE for possible
publication. Copyright may be transferred without notice, after which this
version may no longer be accessible.}
}


\author{\IEEEauthorblockN{Jonas Leo M\"uller\textsuperscript{1,2,*}, Lukas Engel\textsuperscript{3}, Eva Dorschky\textsuperscript{1}, Daniel Krauss\textsuperscript{1,2},\\ Ingrid Ullmann\textsuperscript{3}, Martin Vossiek\textsuperscript{3}, Bj\"orn M. Eskofier\textsuperscript{1,2}}
\IEEEauthorblockA{\textsuperscript{1}Machine Learning and Data Analytics Lab, Friedrich-Alexander-Universit\"at Erlangen-N\"urnberg, Germany\\
\textsuperscript{2}Munich Center for Machine Learning, Germany\\
\textsuperscript{3}Institute of Microwaves and Photonics, Friedrich-Alexander-Universit\"at Erlangen-N\"urnberg, Germany\\
\textsuperscript{*}Corresponding author: jonas.leo.mueller@fau.de}
}

\maketitle

\begin{abstract}
Radar-based human pose estimation enables privacy-preserving motion tracking for ambient intelligence, yet the noisy nature of radar sensing makes uncertainty quantification essential. We present RadProPoser, an end-to-end probabilistic framework that predicts three-dimensional body joints with per-joint uncertainties from raw radar tensor data. Using a variational encoder-decoder with spectral attention that fuses real and imaginary radar components across temporal frames, we model aleatoric uncertainty through learnable Gaussian and Laplace distributions. Trained on a new benchmark dataset with optical motion-capture ground truth, our method achieves 6.425 cm mean per-joint position error. The model outputs per-joint aleatoric uncertainties, and isotonic recalibration yields calibrated total uncertainty with expected calibration error of 0.027. Since spectral attention operates on individual radar tensor components, extending to multi-radar configurations requires only concatenating additional input streams. On the HuPR benchmark with dual orthogonal radars, this achieves 5.042 cm MPJPE. The framework runs at $\sim$89 frames per second (FPS) on an NVIDIA RTX 3090, exceeding the 15 Hz radar frame rate.
\end{abstract}

\begin{IEEEkeywords}
radar-based human pose estimation, variational inference, probabilistic modeling, uncertainty quantification, smart environments
\end{IEEEkeywords}

\section{Introduction}
\label{sec:intro}

Ambient intelligence envisions environments that adapt to human presence, providing context-aware services for healthcare applications such as rehabilitation monitoring, elderly care, and fall detection~\cite{jin2020mmfall, krauss2024review}. Radar-based sensing offers unique advantages for this domain. Unlike cameras, radar operates independently of lighting, functions through occlusions, and preserves privacy by capturing motion signatures rather than identifiable imagery~\cite{ho2024rt, lee2023hupr}. However, radar measurements are inherently stochastic, affected by thermal noise, phase instabilities~\cite{thurn2013noise}, and multipath propagation~\cite{richards2005fundamentals}, making uncertainty quantification essential. Knowing \emph{when} a prediction can be trusted is as important as the prediction itself~\cite{abdar2021}.

\begin{figure*}[htbp]
    \centering
    \includegraphics[width=\textwidth]{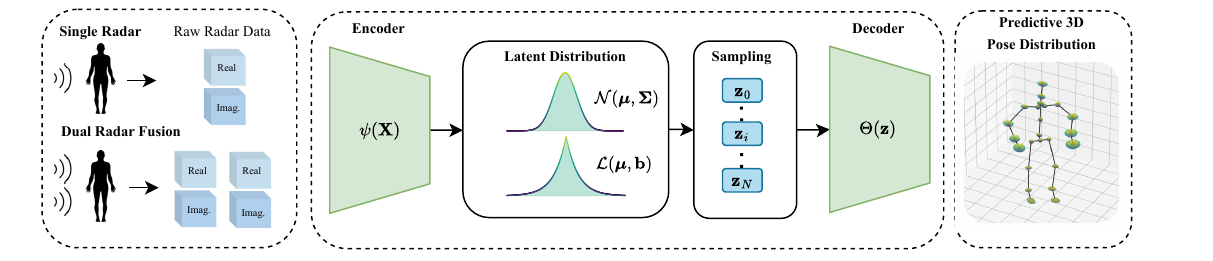}
    \caption{Overview of our approach. An efficient 3D convolutional encoder $\psi(\cdot)$ maps raw radar data $\mathbf{X}$ to a distribution over latent codes. Samples from this distribution are decoded in parallel by $\Theta(\cdot)$ to yield a predictive distribution over 3D poses. We evaluate both Gaussian and Laplace latent priors. Right: predicted pose means with uncertainty visualized as hulls spanning one standard deviation along each axis.}
    \label{fig:preProc}
\end{figure*}

Prior radar-based human pose estimation (HPE) has pursued two approaches, point cloud methods using Constant False Alarm Rate (CFAR) detection~\cite{skolnik1980introduction, sengupta2020nlp}, and radar tensor approaches (RTA) that learn end-to-end from raw measurements~\cite{ho2024rt, lee2023hupr, faucris.330284198}. However, existing RTA benchmarks rely on RGB-based ground truth, introducing uncertainty from prediction errors and depth estimation~\cite{zhang2022survey} that confounds radar-specific uncertainty measurement. Furthermore, prior work has focused on keypoint accuracy alone with limited attention to explicit probabilistic uncertainty modeling~\cite{zheng2022recovering, fan2025diffusion}.

We present RadProPoser (Fig.~\ref{fig:preProc}), a framework for uncertainty quantification in radar-based pose estimation. We introduce a benchmark dataset pairing raw radar measurements with optical motion capture ground truth enabled by a synchronized hardware trigger, ensuring uncertainties reflect the radar sensing process itself rather than image-based preprocessing artifacts. Building on Ho et al.~\cite{ho2024rt}, we extend the backbone with a spectral attention mechanism in complex frequency space that fuses real and imaginary radar components across temporal frames. Since spectral attention operates on individual radar tensor components, extending to multi-radar configurations \cite{lee2023hupr} requires only concatenating additional input streams. Predictive uncertainty comprises two components~\cite{kendall2017uncertainties}. Aleatoric uncertainty is the irreducible ambiguity inherent in the data, here radar noise and multipath effects. Epistemic uncertainty is the reducible uncertainty from limited training data or model capacity. Using variational inference, we learn distributions over latent pose representations, where sampling and decoding multiple latent codes yields predictive distributions whose variance captures aleatoric uncertainty, which dominates in radar sensing where measurements are stochastic. We focus on aleatoric uncertainty modeling as it reflects fundamental hardware limitations (thermal noise, phase instabilities, multipath propagation) that persist regardless of dataset size, making these estimates actionable for system design and sensor placement. We extend these estimates to calibrated total uncertainty through isotonic recalibration~\cite{kuleshov2018accurate} on held-out data. 

Our contributions:
\begin{itemize}
    \item A systematic uncertainty quantification framework for radar tensor-based HPE with our new benchmark dataset.
    \item A spectral attention mechanism for fusing complex radar components that extends to multi-radar configurations, achieving 6.425\,cm MPJPE (single-radar) and 5.042\,cm MPJPE (dual-radar).
    \item Calibrated per-joint uncertainty estimates (ECE of 0.027).
\end{itemize}

Our code and dataset are available at \url{https://github.com/jonasmueler/RadProPoser_IJCNN} and \url{https://zenodo.org/records/14738837}.

\section{Background and Related Work} \label{relatedWork}

\subsection{Radar-based Human Pose Estimation}
Radar-based HPE approaches vary in preprocessing complexity, from raw analog-to-digital converter (ADC) data~\cite{zhao2023cubelearn} to magnitude-based Fast Fourier Transform (FFT) slicing~\cite{lee2023hupr} to compressed point clouds~\cite{sengupta2020mm}, with recent work integrating learnable signal processing~\cite{giroux2023t, chen2025cpformer}. Methods also differ by supervision. Lee et al.~\cite{lee2023hupr} use image-space ground truth with 2D-to-3D lifting~\cite{zhu2023motionbert}, achieving 6.82\,cm MPJPE, while Ho et al.~\cite{ho2024rt} use LiDAR/RGB-derived 3D ground truth with root joint classification and offset regression, achieving 9.91\,cm. Most methods favor direct 3D regression from compressed tensors~\cite{cao2022joint, cao2023task}. We adopt end-to-end regression from FFT-processed complex-valued tensors (Table~\ref{tab:radar_datasets}), ensuring uncertainty estimates reflect data characteristics rather than upstream processing artifacts.

\subsection{Uncertainty Modeling in HPE}
Vision-based HPE research has progressed along two directions. \textit{Uncertainty-aware methods} use uncertainty during training~\cite{zhang2022uncertainty, kundu2022uncertainty} or generative techniques~\cite{gong2023diffpose, feng2023diffpose}. In the radar domain, Chiang et al.~\cite{chiang2024enhancing} apply uncertainty-aware training to point cloud-based pose estimation. \textit{Uncertainty quantification methods}, by contrast, explicitly model predictive uncertainties at inference. The latter includes normalizing flows~\cite{li2021human, dwivedi2024poco}, angular distributions~\cite{prokudin2018deep}, and evidential regression~\cite{bramlage2023plausible}. Direct aleatoric uncertainty regression via Gaussian~\cite{kendall2017uncertainties} or Laplacian~\cite{li2021human} distributions has proven effective. Despite this progress, uncertainty quantification for RTA-based HPE remains underexplored.

\begin{table*}[b]
\small
\setlength{\tabcolsep}{4pt}
\centering
\caption{Raw Tensor-Based Radar Pose Estimation Datasets}
\label{tab:radar_datasets}
\resizebox{\textwidth}{!}{%
\begin{tabular}{l l l l c c l l}
\toprule
\textbf{Authors} & \textbf{Ground Truth} & \textbf{Radar Config.} & \textbf{Evaluation} & \textbf{2D/3D} & \textbf{Keypoints} & \textbf{Method} & \textbf{Reported Score} \\
\midrule
Ho et al.~\cite{ho2024rt} & RGB, LiDAR & 12 TX, 16 RX & MPJPE & 3D & 15 & 3D-HR-layers & 9.91\,cm \\
Lee et al.~\cite{lee2023hupr} & RGB & 2 $\times$ 3 TX, 4 RX & MPJPE & 2D & 14 & Multi-stage feature fusion & 6.82\,cm \\
\midrule
Ours & OMC & 3 TX, 4 RX & MPJPE & 3D & 26 & Prob. modeling & 6.425\,cm \\
\bottomrule
\end{tabular}%
}
\begin{flushleft}
\small MPJPE: Mean per-joint position error; TX: Transmitter; RX: Receiver; OMC: Optical motion capture.
\end{flushleft}
\end{table*}

\section{Methods}
\label{methods}

\subsection{Datasets}

\begin{figure}[htbp]
    \centering
    \begin{minipage}[b]{0.5\columnwidth}
        \centering
        \includegraphics[width=\textwidth]{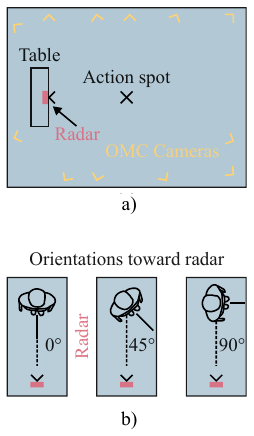}
    \end{minipage}
    \hfill
    \begin{minipage}[b]{0.46\columnwidth}
        \centering
        \includegraphics[width=0.85\textwidth]{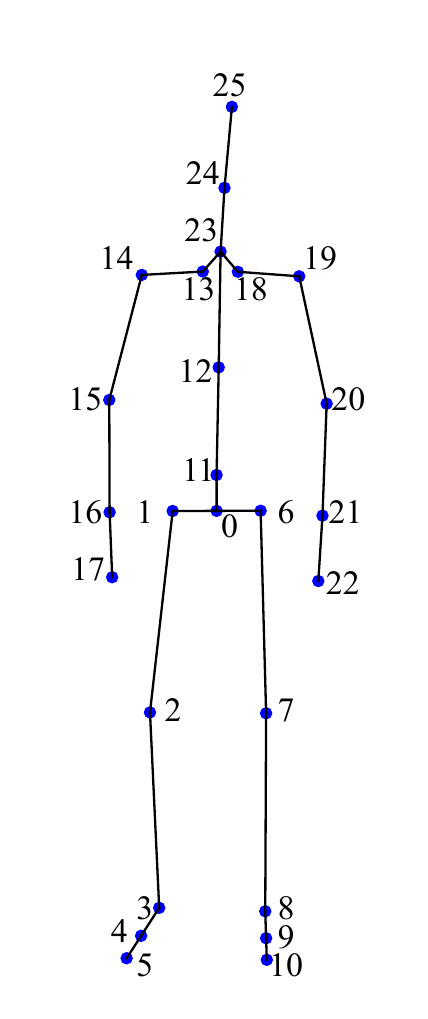}
    \end{minipage}
    \caption{Left: Recording space. Right: Skeleton keypoint structure predicted by RadProPoser.}
    \label{fig:dataCol}
\end{figure}

Our dataset, recorded using the setup of Engel et al.~\cite{Engel2025}, contains radar recordings time-synchronized with optical motion capture (OMC) from 12 participants performing nine exercises at three angles ($0^{\circ}$, $45^{\circ}$, $90^{\circ}$). The exercises include left/right/bilateral upper limb extension, bicep curls, front arm rotation, trunk forward bending, left/right front lunge, and squats. We use a Texas Instruments IWR6843AOPEVM 60\,GHz radar (Table~\ref{tab:radar_parameters}) with ground truth from an Optitrack Flex 13 system.

\begin{table}[htbp]
\centering
\caption{Radar System Parameter Settings}
\label{tab:radar_parameters}
\begin{tabular}{ll}
\toprule
\textbf{Parameter} & \textbf{Value} \\
\midrule
Frequency                      & 60\,GHz \\
Radio-frequency bandwidth      & $\approx$\,1.02\,GHz \\
Frame rate                     & 15\,Hz \\
Chirp duration                 & $\approx$\,17\,$\mu$s \\
Samples per chirp              & 64 \\
ADC sampling frequency         & 3.8\,MHz \\
Chirps per frame               & 128 \\
Number of transmitter antennas & 3 \\
Number of receiver antennas    & 4 \\
TDM-MIMO virtual data channels & 12 \\
\bottomrule
\end{tabular}
\begin{flushleft}
\small ADC: analog-to-digital converter; TDM: time-division multiplexing; MIMO: multiple-input multiple-output.
\end{flushleft}
\end{table}

RadProPoser predicts 26 3D keypoints (Fig.~\ref{fig:dataCol}). Data is split by participant with eight for training, one for hyperparameter tuning (p3), and three for testing (p1, p2, p12). After hyperparameter tuning, models are retrained on all nine non-test participants. Following standard recalibration procedure~\cite{kuleshov2018accurate}, we fit isotonic recalibration on a separate test participant (p1) and evaluate on the remaining two (p2, p12), avoiding data leakage by keeping hyperparameter tuning (p3) and recalibration fitting disjoint.

To demonstrate multi-radar fusion, we evaluate on HuPR~\cite{lee2023hupr}, which uses two orthogonal Texas Instruments IWR1443 radars with only 2D image-space annotations. The original HuPR method predicts 2D keypoints from radar, then lifts to 3D using VideoPose3D~\cite{pavllo20193d}. Since no code is provided for 3D ground truth or prediction, we apply MotionBERT~\cite{zhu2023motionbert}, a 2D-to-3D lifting model achieving ~1\,cm lower MPJPE than VideoPose3D on Human3.6M, to both the 2D annotations and HuPR's released 2D predictions, ensuring identical 3D ground truth for fair comparison. We convert HuPR's 14 keypoints to MotionBERT's 17-joint Human3.6M format via direct mapping and interpolation of pelvis, spine, and thorax. We apply our complex-valued preprocessing (Section~\ref{Reduction}) and train RadProPoser end-to-end using the original train/val/test split.

\subsection{RadProPoser}
\begin{figure*}[htbp]
    \centering
    \includegraphics[width = \textwidth]{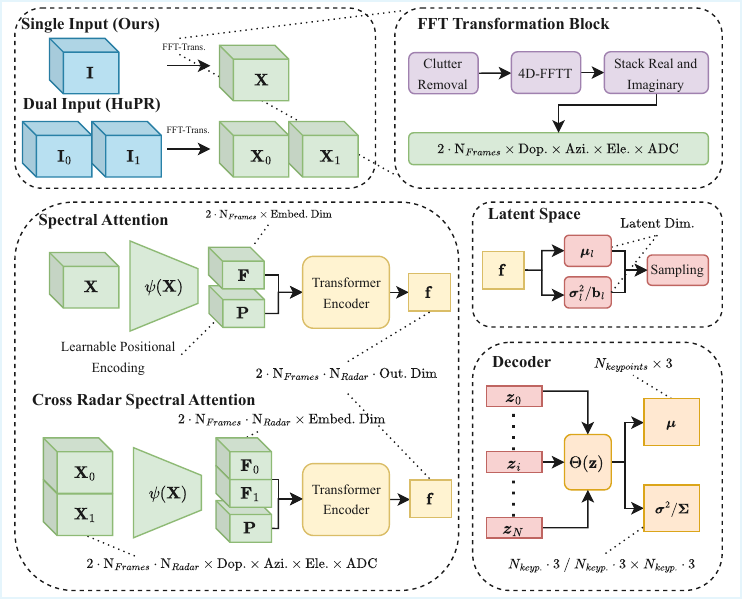}
    \caption{Overview of the RadProPoser architecture. Top left: input handling for single-radar (ours) and dual-radar (HuPR) configurations. Top right: FFT transformation block with clutter removal. Middle left: spectral attention mechanism with learnable positional encodings. Bottom left: cross-radar spectral attention for multi-sensor fusion. Right: latent space parameterization and decoder producing pose predictions with uncertainty estimates.}
    \label{fig:model}
\end{figure*}

The \textbf{RAD}ar \textbf{PRO}babilistic \textbf{POSER} model (RPP) architecture (Fig. \ref{fig:model}) is a function $\phi(\mathbf{I})$ that maps the raw complex-valued radar cube $\mathbf{I} \in \mathbb{C}^{N_{\text{Frames}} \times \text{ADC} \times \text{chirps} \times \text{azimuth} \times \text{elevation}}$ into a distribution over $K$ 3D pose coordinates $\tilde{\boldsymbol{y}} \in \mathbb{R}^{K \times 3}$ using a deterministic encoder and probabilistic decoder, where $K=26$ for our dataset and $K=17$ for HuPR~\cite{lee2023hupr}. This design balances computational efficiency with probabilistic modeling.

\subsubsection{Preprocessing Block}
\label{Reduction}
The raw radar cube $\mathbf{I}$ contains complex-valued samples organized by ADC samples, chirps, and spatial antenna axes (Table~\ref{tab:radar_parameters}). Static clutter is removed by centering the complex radar cube in the chirp dimension, followed by a 4D FFT. This yields $\mathbf{X} \in \mathbb{R}^{2 \cdot N_{\text{Frames}} \times \text{Doppler} \times \text{azimuth} \times \text{elevation} \times \text{ADC}}$, where real and imaginary components are stacked along the frame dimension. We use $N_{\text{Frames}}=8$ (Table~\ref{tab:param_sensitivity}). For multi-radar configurations (e.g., HuPR~\cite{lee2023hupr}), each radar produces its own tensor $\mathbf{X}_i$, stacked to form $\mathbf{X} \in \mathbb{R}^{2 \cdot N_{\text{Radars}} \cdot N_{\text{Frames}} \times \text{Doppler} \times \text{azimuth} \times \text{elevation} \times \text{ADC}}$ for parallel encoding.

\subsubsection{Encoder}
\label{encoder}
The encoder $\psi(\cdot)$ maps $\mathbf{X}$ to frame-wise features for real and imaginary components through a shared 3D Convolutional Neural Network (CNN) (Fig.~\ref{fig:model}), using the 3D HRNet backbone from Ho et al.~\cite{ho2024rt, wang2020deep, cheng2020higherhrnet}. Features from all resolution branches are upsampled, concatenated, and compressed via two 3D conv layers with batch normalization and ReLU, producing $\mathbf{F} \in \mathbb{R}^{2 \cdot N_{\text{Frames}} \times \text{Embed. Dim}}$.

\subsubsection{Spectral Attention}
\label{spectral_attention}
To capture temporal and real/imaginary dependencies, we introduce spectral attention with learnable positional encodings. We implement the attention layer in a standard way with residual connection and MLP. Frame features are projected and positional encodings $\mathbf{P} \in \mathbb{R}^{2 \cdot N_{\text{Frames}} \times d_k}$ are added to queries and keys:
\begin{equation}
    \mathbf{Q} = \mathbf{F}\mathbf{W}_Q + \mathbf{P}, \quad
    \mathbf{K} = \mathbf{F}\mathbf{W}_K + \mathbf{P}, \quad
    \mathbf{V} = \mathbf{F}\mathbf{W}_V,
\end{equation}
where $\mathbf{W}_Q$, $\mathbf{W}_K$, $\mathbf{W}_V$ are learnable projections. The learned encodings project features from different time steps and signal components (real/imaginary) into distinct embedding regions. The attention output is computed as
\begin{equation}
    \text{Attention}(\mathbf{Q}, \mathbf{K}, \mathbf{V}) =
    \text{softmax}\left(\frac{\mathbf{Q}\mathbf{K}^\top}{\sqrt{d_k}}\right)\mathbf{V},
\end{equation}
where $d_k$ is the key dimensionality, enabling each frame to aggregate information across all others while integrating real and imaginary channels.

For multi-radar configurations, features from each radar are concatenated along the frame dimension, yielding $\mathbf{F} \in \mathbb{R}^{2 \cdot N_{\text{Radars}} \cdot N_{\text{Frames}} \times \text{Embed. Dim}}$, with positional encodings $\mathbf{P} \in \mathbb{R}^{2 \cdot N_{\text{Radars}} \cdot N_{\text{Frames}} \times d_k}$ that project each radar into separate embedding regions while enabling cross-radar and temporal fusion. The output $\mathbf{f}$ is obtained by flattening the attention layer output and linear projection.

\subsubsection{Latent Space}
We use variational inference to model a distribution over latent variables conditioned on the radar input, evaluating both isotropic Gaussian and Laplace priors. The spectral attention output $\mathbf{f}$ is projected to produce the latent distribution parameters $(\boldsymbol{\mu}_l, \boldsymbol{\sigma}^2_l)$ for Gaussian or $(\boldsymbol{\mu}_l, \boldsymbol{b}_l)$ for Laplace priors, where $\mathbf{z} \in \mathbb{R}^{256}$ for single-radar and $\mathbf{z} \in \mathbb{R}^{512}$ for dual-radar setups.

For the Gaussian prior, we regularize with Kullback-Leibler (KL) divergence to $\mathcal{N}(\mathbf{0}, \mathbf{I})$
\begin{equation}
\text{KL}_{\mathcal{N}}(\boldsymbol{\mu}_l, \boldsymbol{\sigma}^2_l) = -\frac{1}{2} \left(1 + \log \boldsymbol{\sigma}^2_l - \boldsymbol{\mu}^2_l - \boldsymbol{\sigma}^2_l \right).
\end{equation}
A learnable scaling factor $\alpha$ adjusts sampling variance for downstream tasks. Latent vectors are sampled as
\begin{equation}
\boldsymbol{z}_i = \boldsymbol{\mu}_l + \alpha \boldsymbol{\epsilon}_i \boldsymbol{\sigma}_l, \quad \boldsymbol{\epsilon}_i \sim \mathcal{N}(0, \mathbf{I}).
\end{equation}

For the Laplace latent, we exploit its sharp and well-behaved form for inverse cumulative distribution function sampling without KL regularization~\cite{Habernal.2022.ACL}, sampling as $\boldsymbol{z}_i \sim \mathcal{L}(\boldsymbol{\mu}_l, \boldsymbol{b}_l)$ where $\boldsymbol{b}_l = \text{softplus}(\tilde{\boldsymbol{b}}_l)$ ensures positivity.

\subsubsection{Decoder}
The decoder computes the output distribution
\begin{equation}
\Theta(\boldsymbol{z}) =  \tilde{\boldsymbol{y}},
\end{equation}
where $\Theta$ is a two-layer linear network. We draw $N=500$ samples from the latent distribution, yielding $\boldsymbol{z} \in \mathbb{R}^{N \times \text{Latent Dim.}}$, which the decoder maps to $\tilde{\boldsymbol{y}} \in \mathbb{R}^{N \times K \cdot 3}$. From these we compute mean $\boldsymbol{\mu}$ and variance $\boldsymbol{\sigma}^2 \in \mathbb{R}^{K \cdot 3}$ (or covariance $\boldsymbol{\Sigma}$ when modeling correlations). This keeps the encoder deterministic while enabling parallel sampling via the batch dimension through the probabilistic decoder.

\subsection{Aleatoric Loss Formulations}
We model heteroscedastic aleatoric uncertainty~\cite{kendall2017uncertainties} by minimizing negative log-likelihood (NLL), comparing predicted mean $\boldsymbol{\mu}$ against ground truth $\boldsymbol{y} \in \mathbb{R}^d$, weighted by predicted uncertainty.

The full training loss combines the NLL with a KL divergence term from the variational latent space:
\begin{equation}
\mathcal{L} = \mathrm{NLL}(\boldsymbol{y}, \boldsymbol{\mu}, \boldsymbol{\sigma}^2) + \beta \, \mathrm{KL}(\boldsymbol{\mu}_l, \boldsymbol{\sigma}^2_l),
\end{equation}
where $\beta$ weights the KL regularization.

\paragraph{Gaussian.} Following Kendall and Gal~\cite{kendall2017uncertainties}, we assume independence across output dimensions:
\begin{equation}
\mathrm{NLL}_{\mathrm{Gauss.}}(\boldsymbol{y}, \boldsymbol{\mu}, \boldsymbol{\sigma}^2)
= \sum_{i=1}^{d} \left[ \frac{(\boldsymbol{y}_i - \boldsymbol{\mu}_i)^2}{\boldsymbol{\sigma}^2_i} + \gamma \log \boldsymbol{\sigma}^2_i \right].
\end{equation}

\paragraph{Full Covariance Gaussian.} To capture output correlations, we use full covariance following Gundavarapu et al.~\cite{gundavarapu2019structured}:
\begin{equation}
\mathrm{NLL}_{\mathrm{Cov.}}(\boldsymbol{y}, \boldsymbol{\mu}, \boldsymbol{\Sigma})
= \left(\boldsymbol{y} - \boldsymbol{\mu}\right)^{\top} \boldsymbol{\Sigma}^{-1} \left(\boldsymbol{y} - \boldsymbol{\mu}\right) + \gamma \log\left|\boldsymbol{\Sigma}\right|.
\end{equation}

\paragraph{Laplace.} As an alternative to Gaussian assumptions, we consider independent Laplace likelihoods with scale parameters $\boldsymbol{b} \in \mathbb{R}^d$:
\begin{equation}
\mathrm{NLL}_{\mathrm{Lap.}}(\boldsymbol{y}, \boldsymbol{\mu}, \boldsymbol{b})
= \sum_{i=1}^{d} \left[ \frac{|\boldsymbol{y}_i - \boldsymbol{\mu}_i|}{\boldsymbol{b}_i} + \gamma \log \boldsymbol{b}_i \right].
\end{equation}
This may better capture heavy-tailed errors. In all formulations, $\gamma$ scales the log-variance term ($\gamma=1$ recovers standard NLL). All hyperparameters are provided in the code repository.

\begin{table*}[!t]
\centering
\caption{Average mean per-joint position error $\downarrow$ of three test-set participants for the three recorded angles in cm.}
\label{tab:mpjpe_model_a}
\resizebox{\textwidth}{!} {%
\begin{tabular}{l | ccc | ccc | ccc | ccc | c}
\toprule
\textbf{Model} & \multicolumn{3}{c|}{\textbf{MPJPE (0\degree)}} & \multicolumn{3}{c|}{\textbf{MPJPE (45\degree)}} & \multicolumn{3}{c|}{\textbf{MPJPE (90\degree)}} & \multicolumn{3}{c|}{\textbf{Avg. MPJPE}} & \textbf{Overall Avg.} \\
\hline
& \textbf{P12} & \textbf{P2} & \textbf{P1} &  \textbf{P12} & \textbf{P2} & \textbf{P1} &  \textbf{P12} & \textbf{P2} & \textbf{P1} &  \textbf{P12} & \textbf{P2} & \textbf{P1} \\
\hline
\multicolumn{14}{l}{\textbf{Baselines}} \\
\hline
Ho et al.~\cite{ho2024rt} & 10.881 & 8.700 & 5.946 & 6.476 & 6.208 & 7.438 & 6.481 & 6.544 & 7.945 & 7.946 & 7.151 & 7.110 & 7.402 \\
Engel et al.~\cite{Engel2025}\dag & 11.527 & 8.318 & 6.643 & 6.108 & 6.829 & 7.662 & 6.733 & 7.503 & 9.695 & 8.186 & 7.609 & 8.031 & 7.946 \\
Evidential Regression
& 13.522 & 12.732 & 11.285
& 9.632 & 11.396 & 11.086
& 9.923 & 11.562 & 11.396
& 11.026 & 11.897 & 11.256 & 11.393 \\

\hline
\multicolumn{14}{l}{\textbf{RPP Variants (ours)}} \\
\hline

RPP-Normalizing-Flows
& \textbf{7.417} & 7.632 & 7.097
& 5.788 & 6.085 & 7.023
& 6.639 & 6.941 & 7.196
& 6.947 & 6.886 & 7.072 & 7.022 \\
RPP-Gauss.-Gauss.-Cov.
& 9.369 & \textbf{7.228} & \textbf{5.027}
& \textbf{4.922} & 5.849 & \textbf{6.264}
& 5.635 & 6.307 & 7.220
& \textbf{6.642} & \textbf{6.461} & \textbf{6.170}
& \textbf{6.425} \\
RPP-Gauss.-Gauss.
& 9.970         & 8.028          & 5.127
& 5.556 & 5.710 & 6.374
& \textbf{5.581} & \textbf{6.001} & \textbf{7.022}
& 7.036 & 6.580 & 6.174 & 6.597 \\
RPP-Laplace-Laplace
& 9.988 & 9.061 & 7.783
& 5.591 & 6.665 & 7.098
& 5.683 & 6.949 & 7.321
& 7.087 & 7.558 & 7.401 & 7.349 \\
RPP-Laplace-Gauss.
& 9.063 & 7.641 & 5.203
                    & 5.202 & 5.975 & 6.583
                    & 5.763 & 6.568 & 7.166
                    & 6.676 & 6.728 & 6.317
                    & 6.688 \\
RPP-Gauss.-Laplace
& 10.777 & 9.491 & 7.014
& 5.813 & 7.015 & 7.320
& 6.007 & 7.163 & 8.126
& 7.532 & 7.890 & 7.487 & 7.592 \\
\bottomrule
\end{tabular}%
}
\begin{flushleft}
\small The Models are presented as RPP-[Latent Prior]-[Log-likelihood], with -Cov. indicating full covariance output. \dag Pointcloud-based approach.
\end{flushleft}

\vspace{1em}

\centering
\caption{Procrustes aligned mean per-joint position error $\downarrow$ of three test-set participants for the three recorded angles in cm.}
\label{tab:pmpjpe_model_a}
\resizebox{\textwidth}{!} {%
\begin{tabular}{l | ccc | ccc | ccc | ccc | c}
\toprule
\textbf{Model} & \multicolumn{3}{c|}{\textbf{P-MPJPE (0\degree)}} & \multicolumn{3}{c|}{\textbf{P-MPJPE (45\degree)}} & \multicolumn{3}{c|}{\textbf{P-MPJPE (90\degree)}} & \multicolumn{3}{c|}{\textbf{Avg. P-MPJPE}} & \textbf{Overall Avg.} \\
\hline
& \textbf{P12} & \textbf{P2} & \textbf{P1} &  \textbf{P12} & \textbf{P2} & \textbf{P1} &  \textbf{P12} & \textbf{P2} & \textbf{P1} &  \textbf{P12} & \textbf{P2} & \textbf{P1} \\
\hline
\multicolumn{14}{l}{\textbf{Baselines}} \\
\hline
Ho et al.~\cite{ho2024rt} & 5.939 & 6.048 & 4.524
                       & 4.249 & 4.540 & 5.559
                       & 5.097 & 5.427 & 6.315
                       & 5.095 & 5.339 & 5.466
                       & 5.300 \\
Engel et al.~\cite{Engel2025}\dag &  \textbf{5.122} & \textbf{5.384} & 4.521
                       & 4.132 & 4.638 & 5.625
                       & 5.156 & 5.797 & 7.762
                       & 4.803 & 5.273 & 5.969
                       & 5.348 \\
Evidential Regression
                       & 10.417 & 12.130 & 9.576
                       & 9.346 & 11.718 & 10.394
                       & 9.989 & 12.118 & 10.709
                       & 9.917 & 11.989 & 10.226
                       & 10.711 \\

\hline
\multicolumn{14}{l}{\textbf{RPP Variants (ours)}} \\
\hline

RPP-Normalizing-Flows
& 6.264 & 7.028 & 5.168
& 4.363 & 5.503 & 6.424
& 5.368 & 6.082 & 6.725
& 5.665 & 6.204 & 6.106 & 6.325 \\

RPP-Gauss.-Gauss.  & 5.939 & 6.389 & 4.764
                       & 3.948 & 4.608 & 5.290
                       & 4.900 & 5.423 & \textbf{5.793}
                       & 4.929 & 5.473 & 5.282
                       & 5.228 \\
RPP-Gauss.-Gauss.-Cov. &
5.363 & 5.516 & \textbf{3.970}
& \textbf{3.738} & \textbf{4.128} & \textbf{5.074}
& \textbf{4.654} & \textbf{4.797} & 5.918
& \textbf{4.585} & \textbf{4.813} & \textbf{4.987}
& \textbf{4.795} \\
RPP-Laplace-Laplace
                       & 6.601 & 7.396 & 5.116
                       & 4.374 & 5.152 & 5.807
                       & 4.870 & 5.938 & 6.287
                       & 5.282 & 6.162 & 5.737
                       & 5.727 \\
RPP-Laplace-Gauss.
                       & 5.772 & 6.050 & 4.237
                & 3.980 & 4.250 & 5.416
                & 4.865 & 5.377 & 5.942
                & 4.872 & 5.226 & 5.198
                & 5.099 \\
RPP-Gauss.-Laplace
                       & 6.927 & 7.950 & 5.131
                       & 4.692 & 5.680 & 6.155
                       & 5.238 & 6.186 & 6.841
                       & 5.619 & 6.605 & 6.042
                       & 6.200 \\

\bottomrule
\end{tabular}%
}
\begin{flushleft}
\small The Models are presented as RPP-[Latent Prior]-[Log-likelihood], with -Cov. indicating full covariance output. \dag Pointcloud-based approach.
\end{flushleft}
\end{table*}

\subsection{Calibration and Recalibration}
Probabilistic models often produce overconfident or underconfident predictions~\cite{kuleshov2018accurate,bramlage2023plausible}. A distributional predictor is calibrated if predicted confidence levels match empirical frequencies. We quantify miscalibration using expected calibration error (ECE)~\cite{kuleshov2018accurate}
\begin{equation}
\text{ECE} = \frac{1}{C} \sum_{j=1}^C \left| \hat{P}(p_j) - p_j \right|,
\end{equation}
where $C$ is the number of confidence levels and $\hat{P}(p_j)$ is the empirical frequency of samples below target coverage level $p_j$. We also evaluate sharpness, the mean predicted variance
\begin{equation}
\text{Sharpness} = \frac{1}{NK} \sum_{k=1}^K \sum_{i=1}^N \sigma^2_{i, k},
\end{equation}
since well-calibrated but overly wide intervals provide little actionable information. To correct miscalibration, we apply isotonic regression to learn a recalibration mapping $R : [0,1] \to [0,1]$ between the model's predicted cumulative distribution function (CDF) values and empirical coverage probabilities~\cite{kuleshov2018accurate}. This non-parametric method enforces monotonicity while fitting the empirical calibration curve. The learned mapping transforms the predicted distributions such that recalibrated confidence levels accurately reflect empirical coverage rates.

\section{Experiments} \label{experiments}
\subsection{Experimental Setup}
We evaluate six RadProPoser (RPP) variants combining two latent spaces (Gaussian, Laplace) with three output likelihoods (Gaussian diagonal, Gaussian full covariance, Laplace). The Laplace latent space is motivated by Li et al.~\cite{li2021human}, who showed heavier-tailed distributions improve pose estimation. We additionally evaluate a normalizing flow latent space~\cite{rezende2015variational} using RealNVP~\cite{dinh2017realnvp}, which captures total uncertainty directly without aleatoric decomposition.

As probabilistic baseline, evidential regression~\cite{amini2020deep} provides closed-form aleatoric uncertainty without sampling. As non-probabilistic baselines, we adapt the single-frame HRNet encoder from Ho et al.~\cite{ho2024rt} and retrain the point cloud model from Engel et al.~\cite{Engel2025} on our split. Lee et al.~\cite{lee2023hupr} are excluded from direct comparison due to incompatible dual-radar 2D predictions, but we evaluate on HuPR separately.

Following Zheng et al.~\cite{zheng2023deep}, we measure pose accuracy using Mean Per-Joint Position Error (MPJPE) and Procrustes-aligned MPJPE (P-MPJPE), which isolates structural accuracy via rigid transformation. For probabilistic models, predictive uncertainty per keypoint is $\boldsymbol{u}_k = \sum_{d=1}^{3} \boldsymbol{\sigma}^2_{k, d}$, with calibration assessed using ECE and sharpness~\cite{kuleshov2018accurate}.

\section{Results and Discussion}

We evaluate RadProPoser across four dimensions. First, we compare all model variants on our dataset for pose estimation accuracy (Tables~\ref{tab:mpjpe_model_a} and~\ref{tab:pmpjpe_model_a}), uncertainty quantification (Table~\ref{tab:calibration_sharpness}), and inference efficiency (Table~\ref{tab:inference_time}). We then evaluate the best-performing configuration (RPP-Gauss.-Gauss.-Cov.) for multi-radar sensor fusion on the HuPR benchmark.

\subsection{Pose Estimation Accuracy}
Our best-performing model, RPP-Gauss.-Gauss.-Cov., achieves 6.425\,cm MPJPE, representing a 13.2\% improvement over the adapted baseline from Ho et al.~\cite{ho2024rt} (7.402\,cm) and a 19.1\% improvement over the point cloud method by Engel et al.~\cite{Engel2025} (7.946\,cm). Under Procrustes alignment, which isolates structural pose accuracy from global positioning, our model achieves 4.795\,cm P-MPJPE compared to 5.300\,cm for Ho et al. and 5.348\,cm for Engel et al., improvements of 9.5\% and 10.3\% respectively.

The full-covariance Gaussian formulation outperforms the diagonal variant (6.425\,cm vs. 6.597\,cm) by modeling correlations between joint positions. The Normalizing Flow model achieves 7.022\,cm, while evidential regression performs worse (11.393\,cm). All multi-frame RadProPoser variants outperform the single-frame baseline from Ho et al.~\cite{ho2024rt}.

\begin{table*}[!t]
\centering
\caption{Pose Estimation Performance on HuPR Test Set (cm) $\downarrow$}
\label{tab:hupr_pose_performance}
\resizebox{\textwidth}{!}{%
\begin{tabular}{l | l | cc | cc | cc | cc | cc}
\toprule
\textbf{Model} & \textbf{GT Source} & \multicolumn{2}{c|}{\textbf{Elbow}} & \multicolumn{2}{c|}{\textbf{Wrist}} & \multicolumn{2}{c|}{\textbf{Knee}} & \multicolumn{2}{c|}{\textbf{Ankle}} & \multicolumn{2}{c}{\textbf{Total}} \\
\cmidrule{3-12}
& & \textbf{MPJPE} & \textbf{P-MPJPE} & \textbf{MPJPE} & \textbf{P-MPJPE} & \textbf{MPJPE} & \textbf{P-MPJPE} & \textbf{MPJPE} & \textbf{P-MPJPE} & \textbf{MPJPE} & \textbf{P-MPJPE} \\
\midrule
mmMesh$^\dagger$ & VideoPose3D & 11.29 & -- & 21.82 & -- & -- & -- & -- & -- & 7.13 & -- \\
HuPR$^{\dagger\ddagger}$ & VideoPose3D & 8.53 & -- & 15.64 & -- & -- & -- & -- & -- & 6.82 & -- \\
HuPR$^\ddagger$ & MotionBERT & 16.883 & 12.755 & 31.292 & 25.696 & 4.979 & 6.354 & 7.226 & 9.189 & 9.610 & 9.233 \\
RPP-Gauss.-Gauss.-Cov. & MotionBERT & \textbf{8.168} & \textbf{6.065} & \textbf{14.612} & \textbf{11.886} & \textbf{3.244} & \textbf{3.056} & \textbf{5.538} & \textbf{5.262} & \textbf{5.042} & \textbf{4.741} \\
\bottomrule
\end{tabular}%
}
\begin{flushleft}
\small $^\dagger$Results taken from original publication. $^\ddagger$Uses 2D-to-3D lifting at inference. GT Source indicates the lifting model used to generate 3D pseudo-ground-truth from HuPR's 2D annotations; RPP predicts 3D poses directly. Per-joint values are averaged over left and right keypoints.
\end{flushleft}
\end{table*}

Table~\ref{tab:ablation} shows spectral attention improves performance on both single and dual radar configurations.

\begin{table}[h]
\centering
\caption{Spectral Attention Ablation}
\label{tab:ablation}
\begin{tabular}{lccc}
\toprule
\textbf{Configuration} & \textbf{Spec. Att.} & \textbf{MPJPE} & \textbf{P-MPJPE} \\
\midrule
Our dataset (single radar) & \texttimes & 7.992 & 5.733 \\
Our dataset (single radar) & \checkmark & 6.425 & 4.795 \\
\midrule
HuPR (dual radar) & \texttimes & 5.999 & 5.833 \\
HuPR (dual radar) & \checkmark & 5.042 & 4.741 \\
\bottomrule
\end{tabular}
\begin{center}
\small MPJPE and P-MPJPE in cm.
\end{center}
\end{table}

Table~\ref{tab:param_sensitivity} shows the impact of frame count and latent dimensionality, where 8 frames and 256-dimensional latent space yield optimal results.

\begin{table}[h]
\centering
\caption{Impact of Frame Count and Latent Dimensionality}
\label{tab:param_sensitivity}
\begin{tabular}{l|ccc|cc}
\toprule
& \multicolumn{3}{c|}{$d = 256$} & \multicolumn{2}{c}{$t = 8$} \\
\midrule
$t$ / $d$ & 2 & 4 & 8 & 512 & 1024 \\
MPJPE (cm) & 7.357 & 7.215 & 6.425 & 6.961 & 7.332 \\
\bottomrule
\end{tabular}
\begin{center}
\small $t = N_{\text{Frames}}$, $d$ = Latent Dim.
\end{center}
\end{table}

\subsection{Multi-Radar Sensor Fusion on HuPR}
\label{sec:hupr_results}
To demonstrate generalization to different radar configurations, we train RPP-Gauss.-Gauss.-Cov. on the HuPR benchmark~\cite{lee2023hupr} with dual orthogonally-mounted radars (Table~\ref{tab:hupr_pose_performance}).
Our model achieves 5.042\,cm MPJPE compared to HuPR's 9.610\,cm with identical MotionBERT~\cite{zhu2023motionbert} lifting and using the given model weights for 2D prediction, a 47.5\% improvement. This gain reflects our spectral attention mechanism effectively fusing multi-radar inputs and our preprocessing preserving full complex-valued signal content. The improvement is particularly pronounced for challenging keypoints with wrist error decreasing from 31.292\,cm to 14.612\,cm (53.3\% reduction) and elbow error from 16.883\,cm to 8.168\,cm (51.6\% reduction), validating our end-to-end approach.

\subsection{Uncertainty Quantification and Calibration}
\begin{table}[h]
\centering
\caption{Calibration and Sharpness on Our Test Set}
\label{tab:calibration_sharpness}
\resizebox{\columnwidth}{!}{%
\begin{tabular}{l | cc | cc}
\toprule
\textbf{Model} & \textbf{ECE} & \textbf{ECE (Cal.)} & \textbf{Sharpness} & \textbf{Sharpness (Cal.)} \\
\midrule
RPP-Gauss.-Gauss.-Cov. & 0.140 & 0.027 & 6.290 & 40.884 \\
RPP-Gauss.-Gauss.      & 0.163 & 0.031 & 2.146 & 16.871 \\
RPP-Laplace-Gauss.     & 0.171 & 0.022 & 1.938 & 12.964 \\
RPP-Gauss.-Laplace     & 0.185 & 0.021 & 1.532 & 14.771 \\
RPP-Laplace-Laplace    & 0.154 & 0.026 & 8.970 & 58.372 \\
RPP-Normalizing-Flows  & 0.078 & 0.044 & 28.623 & 57.061 \\
Evidential Regression  & 0.180 & 0.042 & 9054.15 & 4906.734 \\
\bottomrule
\end{tabular}%
}
\begin{flushleft}
\small RPP models are presented as RPP-[Latent Prior]-[Log-likelihood], with -Cov. indicating full covariance output. Sharpness in cm\textsuperscript{2}. Cal.: calibrated.
\end{flushleft}
\end{table}
Table~\ref{tab:calibration_sharpness} summarizes calibration performance. Prior to recalibration, models exhibit moderate miscalibration with ECE ranging from 0.078 to 0.185. After isotonic recalibration, all models achieve ECE below 0.05, with the best result of ECE = 0.021 (RPP-Gauss.-Laplace). RPP-Gauss.-Gauss.-Cov. achieves ECE = 0.027 while maintaining the highest pose accuracy, offering flexibility between raw aleatoric estimates and calibrated total uncertainty for downstream decision-making. The sharpness values reflect this correction. RPP-Gauss.-Gauss.-Cov. increases from 6.290 to 40.884 cm² as overconfident predictions are relaxed to achieve accurate coverage.
\begin{table}[h]
    \centering
    \caption{MPJPE and Uncertainty by Body Region}
    \resizebox{\columnwidth}{!}{%
    \begin{tabular}{lccccc}
    \toprule
    & \textbf{Center} & \textbf{Left Leg} & \textbf{Right Leg} & \textbf{Left Arm} & \textbf{Right Arm} \\
    & (n=6) & (n=5) & (n=5) & (n=5) & (n=5) \\
    \midrule
    \textbf{MPJPE} & 5.649 & 5.069 & 5.379 & 9.115 & 8.077 \\
    \textbf{Uncal.} & 4.825 & 4.065 & 3.861 & 10.149 & 8.845 \\
    \textbf{Cal.} & 28.526 & 25.606 & 20.847 & 72.318 & 59.596 \\
    \bottomrule
    \end{tabular}%
    }
    \begin{flushleft}
    \small MPJPE in cm, uncertainty (Uncal./Cal.) in cm². Values averaged over keypoints per region. Results for RPP-Gauss.-Gauss.-Cov.
    \end{flushleft}
    \label{tab:keypoints_grouped_calibrated}
\end{table}
Per-keypoint analysis (Table~\ref{tab:keypoints_grouped_calibrated}) shows that learned uncertainties reflect physical signal properties. Extremities with smaller reflective surfaces exhibit higher errors and proportionally higher uncertainties, while torso keypoints show the lowest of both. This approach generalizes to HuPR, where RPP-Gauss.-Gauss.-Cov. improves from ECE 0.103 to 0.023 after recalibration (Table~\ref{tab:hupr_recalibration}). While sharpness increases from 5.991 to 22.371 cm² after recalibration, the uncertainties remain informative. The lower overall sharpness compared to our single-radar setup likely reflects the additional information from dual-radar fusion. Wrist predictions show the highest sharpness (94.554 cm²), consistent with known limitations of 2D-to-3D lifting models~\cite{zhu2023motionbert} for arm estimation. The calibrated uncertainties thus capture errors from both radar sensing and the lifting model used to generate ground truth.
\begin{table}[h]
\centering
\caption{Calibration and Sharpness on HuPR Test Set}
\label{tab:hupr_recalibration}
\resizebox{\columnwidth}{!}{%
\begin{tabular}{lcccccc}
\toprule
\textbf{Method} & \textbf{ECE} $\downarrow$ & \textbf{Elbow} & \textbf{Wrist} & \textbf{Knee} & \textbf{Ankle} & \textbf{Total} \\
\midrule
Uncalibrated & 0.103 & 7.686 & 19.183 & 4.172 & 9.786 & 5.991 \\
Calibrated & 0.023 & 30.718 & 94.554 & 9.607 & 25.952 & 22.371 \\
\bottomrule
\end{tabular}%
}
\begin{flushleft}
\small Sharpness in cm². Per-joint values are averaged over left and right keypoints.
\end{flushleft}
\end{table}
\subsection{Inference Efficiency}
\begin{table}[h]
\centering
\caption{Inference Time Analysis}
\label{tab:inference_time}
\resizebox{\columnwidth}{!}{%
\begin{tabular}{lccc}
\toprule
\textbf{Model} & \textbf{Params (M)} & \textbf{FLOPs (G)} & \textbf{Time (ms)} \\
\midrule
RPP-Gauss.-Gauss.-Cov.      & 9.514 & 37.841 & 11.204 \\
RPP-Normalizing-Flows      & 10.04 & 38.359 & 12.152 \\
Evidential Regression      & 9.019 & 37.791 & 11.048 \\
Ho et al.~\cite{ho2024rt}  & 7.762 & 2.357  & 4.286  \\
\bottomrule
\end{tabular}%
}
\begin{flushleft}
\small Single-radar configuration. Batch size = 1, averaged over 50 repetitions, G = giga, M = million.
\end{flushleft}
\end{table}
RadProPoser achieves real-time performance at approximately 11\,ms per frame ($\sim$89 FPS) on an NVIDIA RTX 3090 GPU, exceeding the 15\,Hz radar frame rate (Table~\ref{tab:inference_time}). The encoder processes input frames in parallel through 3D convolutions. We then draw $N=500$ samples from the latent distribution, stack them in the batch dimension, and process them through the decoder in one parallel forward pass to obtain the full output distribution. The method of Ho et al.~\cite{ho2024rt} is faster (4.286\,ms) due to single-frame processing, but at reduced accuracy.
%
\subsection{Limitations and Future Directions}

Our dataset was recorded in a controlled laboratory environment to ensure reliable uncertainty estimation, while evaluation on HuPR demonstrates generalization across sensor configurations. Future work will explore uncertainty-guided data simulation~\cite{schussler2021realistic}, and employ the calibrated per-joint uncertainties for downstream applications including biomechanical simulations~\cite{wakeling2023review} and digital physiotherapy~\cite{10.1007/978-3-032-00652-3_17}.

\section{Conclusion} \label{conclusion}

This work introduces RadProPoser, the first uncertainty quantification framework for radar-based human pose estimation. Through variational inference with full covariance modeling and spectral attention for multi-radar fusion, RadProPoser provides calibrated per-joint uncertainties that reflect physical signal properties. The framework generalizes across sensor configurations and establishes radar as a viable modality for trustworthy ambient intelligence systems that enhance human wellbeing without compromising privacy.

%

\bibliographystyle{IEEEtran}
\bibliography{bibl}

\end{document}